\documentclass{article}

\usepackage{PRIMEarxiv}

\usepackage[utf8]{inputenc} 
\usepackage[T1]{fontenc}    
\usepackage{hyperref}       
\usepackage{url}            
\usepackage{booktabs}       
\usepackage{amsfonts}       
\usepackage{nicefrac}       
\usepackage{microtype}      
\usepackage{lipsum}
\usepackage{fancyhdr}       
\usepackage{graphicx}       
\usepackage{orcidlink}
\graphicspath{{media/}}     

\usepackage{amsmath}
\usepackage{booktabs}
\usepackage{graphicx}     
\usepackage{tikz}         
\usepackage{pgfplots}     
\pgfplotsset{compat=1.18}
\usepackage{tikz}
\usetikzlibrary{positioning, arrows.meta, arrows.meta}
\usepackage{bbm}
\usepackage{pgfplotstable}
\usepackage{amsmath} 
\usepackage{orcidlink}
\usepackage{pgf-pie} 
\usepackage{xcolor} 

\usepackage{tikz}
\usetikzlibrary{shapes.geometric, arrows, positioning}
\usetikzlibrary{shadows}

\title{BeliefShift: Benchmarking Temporal Belief Consistency and Opinion Drift in LLM Agents}

\author{
  Praveen Kumar Myakala \orcidlink{0009-0009-6988-5592}\\
  Independent AI Researcher, Texas, USA \\
  \texttt{Praveen.K.Myakala@gmail.com}
  \And
  Manan Agrawal \orcidlink{0009-0007-4289-8137}\\
  University of North Carolina at Charlotte, \\ North Carolina, USA \\
  \texttt{mananua25gmail.com}
  \And
  Rahul Manche \orcidlink{0009-0000-1861-4084}\\
  Independent Researcher, Hyderabad, India\\
  \texttt{rahulmanche1@gmail.com}
}

\begin{document}
\maketitle

\begin{abstract}
LLMs are increasingly used as long-running conversational agents, yet every major benchmark evaluating their memory treats user information as static facts to be stored and retrieved. That's the wrong model. People change their minds, and over extended interactions, phenomena like opinion drift, over-alignment, and confirmation bias start to matter a lot.\par\vspace{0.5em}

\textbf{BeliefShift} introduces a longitudinal benchmark designed specifically to evaluate belief dynamics in multi-session LLM interactions. It covers three tracks: \textit{Temporal Belief Consistency}, \textit{Contradiction Detection}, and \textit{Evidence-Driven Revision}. The dataset includes 2,400 human-annotated multi-session interaction trajectories spanning health, politics, personal values, and product preferences.\par\vspace{0.5em}

We evaluate seven models including GPT-4o, Claude 3.5 Sonnet, Gemini 1.5 Pro, LLaMA-3, and Mistral-Large under zero-shot and retrieval-augmented generation (RAG) settings. Results reveal a clear trade-off: models that personalize aggressively resist drift poorly, while factually grounded models miss legitimate belief updates.\par\vspace{0.5em}

We further introduce four novel evaluation metrics: \textit{Belief Revision Accuracy} (BRA), \textit{Drift Coherence Score} (DCS), \textit{Contradiction Resolution Rate} (CRR), and \textit{Evidence Sensitivity Index} (ESI).
\end{abstract}

\section{Introduction}
\label{sec:intro}

Large Language Models (LLMs) have rapidly evolved from single-turn question-answering systems into persistent conversational agents deployed across education, personal productivity, healthcare, and mental wellness applications~\cite{openai2023gpt4, anthropic2024claude}. Users now return to these systems repeatedly, sometimes over months, building up an interaction history that the agent is expected to remember and reason over. This shift from stateless to stateful, long-horizon deployment fundamentally changes what we should demand from these systems~\cite{myakala2025xai}.

Yet our benchmarks have not kept up. The dominant assumption baked into current evaluation frameworks is that user-provided information is static: a fact stated once remains true indefinitely, and the agent's job is simply to retrieve it accurately when needed. Benchmarks like LoCoMo~\cite{maharana2024locomo} and LongMemEval~\cite{wu2024longmemeval} have pushed the field forward on episodic memory retention, but they share this same blind spot. They measure whether an agent remembers \textit{what} a user said. They do not measure whether the agent understands that what a user believes today may not be what they believed three weeks ago.

This is not a corner case. Human beliefs, preferences, and opinions are inherently dynamic. A user's stance on a medical treatment may shift after reading new research. Political views evolve in response to lived experience. Product preferences change. Research has shown that LLM agents can develop digital mirrors of user identity over extended interactions~\cite{methuku2025doppelgangers}, and that accumulated context systematically shifts model behavior in ways that are difficult to detect or control~\cite{geng2025accumulating}. In extended interactions with an LLM agent, these shifts create a set of challenges that no existing benchmark is equipped to evaluate.

\subsection*{Core Challenges}

We identify three under-addressed challenges that motivate this work:

\textbf{Belief change as a first-class citizen.} Current benchmarks treat every user utterance as a ground-truth fact to be stored. When a user contradicts an earlier statement, this is either ignored or flagged as noise. There is no framework for modeling belief change as a meaningful, expected event that the agent should track and represent faithfully~\cite{hase2024belief, jang2024beliefr}.

\textbf{Distinguishing drift from revision.} Not all opinion changes are equal. A user who updates their view after encountering new evidence is engaging in rational belief revision, where the new belief is \textit{caused} by evidence \(\textit{User} + \textit{Evidence} \rightarrow \textit{NewBelief}\). A user whose opinion shifts simply because the model keeps nudging them is experiencing drift, a failure mode where model bias does the work instead \(\textit{User} + \textit{ModelBias} \rightarrow \textit{NewBelief}\). This distinction is operationalized directly in our Evidence Sensitivity Index (ESI) metric. These two phenomena look similar on the surface but have very different implications for agent trustworthiness~\cite{sharma2023sycophancy, dongre2025drift, fanous2025syceval}. No existing metric separates them.

\textbf{Longitudinal interaction corpora.} Evaluating belief dynamics requires dialogue data that captures how user stances evolve across multiple sessions over time, with coherent narrative continuity between sessions. Such corpora do not currently exist at the scale or annotation depth needed for rigorous benchmarking~\cite{maharana2024locomo, wu2024longmemeval}.

\subsection*{BeliefShift}

We introduce \textbf{BeliefShift}, the first longitudinal benchmark explicitly designed to evaluate how well LLM agents track, represent, and reason about evolving user beliefs across extended multi-session interactions. BeliefShift is built around three evaluation tracks:

\begin{itemize}
    \item \textit{Temporal Belief Consistency}: whether agents correctly recall and update a user's changing stance over time.    \item \textit{Contradiction Detection}: whether agents identify and appropriately reconcile contradictory positions expressed across sessions~\cite{ranaldi2023sycophantic}.
    \item \textit{Evidence-Driven Revision}: whether agents distinguish uninformed opinion changes from belief revisions grounded in new factual information~\cite{cheng2025accommodation, borah2025mind}.
\end{itemize}

Critically, BeliefShift provides the first quantitative framework for measuring what prior work has theorized as the \textit{mirroring effect}, the tendency of LLM agents to progressively reflect and reinforce user identity over sustained interactions~\cite{methuku2025doppelgangers, myakala2025xai}.

The benchmark comprises 2,400 multi-session interaction trajectories spanning topics in health, politics, personal values, and product preferences, simulating interactions across 10 to 50 conversational sessions per trajectory. Dialogues combine human-authored scenarios with synthetically generated trajectories, all carrying human-annotated belief state labels at the session level.

To support evaluation, we introduce four novel metrics: \textit{Belief Revision Accuracy} (BRA), \textit{Drift Coherence Score} (DCS), \textit{Contradiction Resolution Rate} (CRR), and \textit{Evidence Sensitivity Index} (ESI). Together, these metrics operationalize a distinction the field has so far lacked: the difference between an agent that adapts intelligently to genuine belief change and one that simply drifts with the user~\cite{dongre2025drift, geng2025accumulating}.

We evaluate seven state-of-the-art LLMs, including GPT-4o~\cite{openai2024gpt4o}, Claude 3.5 Sonnet~\cite{anthropic2024claude}, Gemini 1.5 Pro~\cite{geminiteam2024gemini15}, LLaMA-3~\cite{dubey2024llama3}, and Mistral-Large~\cite{jiang2023mistral}, under both zero-shot and retrieval-augmented generation (RAG)~\cite{lewis2020rag} settings and find a consistent tension across all models: systems that personalize aggressively close opinion gaps quickly but drift badly, while factually grounded models resist drift but fail to track legitimate belief updates. No current architecture handles both well. BeliefShift is designed to make that gap visible, measurable, and impossible to ignore.

\vspace{0.5em}
\noindent The rest of the paper is organized as follows. Section~\ref{sec:related} reviews related work on memory benchmarking, belief modeling, and opinion dynamics in LLMs. Section~\ref{sec:benchmark} describes the BeliefShift benchmark design and dataset construction. Section~\ref{sec:metrics} formally defines the four evaluation metrics. Section~\ref{sec:experiments} presents experimental results, and Section~\ref{sec:discussion} discusses implications and limitations. Section~\ref{sec:conclusion} concludes.

\section{Related Work}
\label{sec:related}

BeliefShift sits at the intersection of several active research threads: long-term conversational memory, sycophancy and alignment failure, belief revision in language models, and opinion dynamics over extended interactions. We review each in turn and position BeliefShift relative to prior work.

\subsection{Episodic Memory Benchmarks}

The evaluation of long-term memory in conversational agents has seen significant progress in recent years. LoCoMo~\cite{maharana2024locomo} introduced a benchmark of 35-session dialogues designed to test factual recall and temporal reasoning over extended interactions, establishing a strong baseline for episodic memory evaluation. LongMemEval~\cite{wu2024longmemeval} extended this line of work by explicitly testing knowledge updates, particularly factual changes such as a user's new address or job title. While both benchmarks represent important advances, they share a fundamental limitation: the user's beliefs, preferences, and opinions are treated as static anchors rather than as dynamic, evolving states. BeliefShift addresses precisely this gap, treating belief change not as noise to be filtered but as a first-class phenomenon to be tracked and evaluated.

\subsection{Sycophancy and Over-Alignment}

A growing body of work has documented the tendency of LLMs to mirror and reinforce user stances rather than maintain independent, well-grounded positions. Sharma et al.~\cite{sharma2023sycophancy} provided the first systematic analysis of sycophantic behavior in LLMs, showing that models trained with reinforcement learning from human feedback (RLHF) are particularly susceptible to validating user opinions regardless of their factual basis. Ranaldi and Pucci~\cite{ranaldi2023sycophantic} further demonstrated that LLMs exhibit contradictory behavior when user pressure conflicts with factual grounding. More recent work by Fanous et al.~\cite{fanous2025syceval} introduced SycEval, a dedicated benchmark for measuring sycophancy across a range of task types. Borah et al.~\cite{borah2025mind} showed that LLMs exhibit amplified belief congruence, progressively mirroring the belief systems of the personas they interact with, a phenomenon closely related to what Methuku and Myakala~\cite{methuku2025doppelgangers} theorized as the mirroring effect in AI identity formation over sustained interactions. 

\begin{figure}[t]
\centering
\begin{tikzpicture}[
  font=\small,
  box/.style={
    rectangle, rounded corners=3pt,
    draw=black!70, very thick,
    fill=#1, text width=3.0cm,
    align=center, minimum height=1.0cm
  },
  loop/.style={->, very thick, >=Stealth, draw=red!65},
  break/.style={->, thick, >=Stealth, dashed, draw=blue!65},
  lbl/.style={font=\footnotesize, inner sep=1pt, fill=white},
  callout/.style={
    rectangle, rounded corners=3pt,
    draw=blue!60, thick,
    fill=blue!6, align=left,
    text width=3.6cm, font=\footnotesize,
    inner sep=6pt
  }
]

\node[box=yellow!18] (user)   at (0,  1.8) {User Belief\\$U_t$};
\node[box=orange!18] (input)  at (5,  1.8) {User Input};

\node[box=red!10]    (model)  at (5, -1.8) {LLM Response};
\node[box=red!16]    (mirror) at (0, -1.8) {Reinforced Belief\\$U_{t+1}$};

\draw[loop] (user)  -- (input)
  node[lbl, pos=0.5, above] {expresses};

\draw[loop] (input) -- (model)
  node[lbl, pos=0.4, right] {sycophantic alignment};

\draw[loop] (model) -- (mirror)
  node[lbl, pos=0.5, below] {reinforces};

\draw[loop] (mirror) -- (user)
  node[lbl, pos=0.5, left] {amplifies $U_t$};

\node[font=\footnotesize, text=red!70] at (2.5, 0.0) {Mirroring (feedback loop)};

\node[callout, anchor=west] (dcsbox) at (8.2, 0.0)
{\textbf{BeliefShift: DCS}\\Detects coherent drift by measuring whether\\successive sessions move in a self-reinforcing\\direction without new grounding.};

\draw[break] (5, 0.0) -- (dcsbox.west);

\end{tikzpicture}
\caption{The sycophancy feedback loop leading to the mirroring effect. Over successive sessions, the model progressively reinforces user beliefs $U_t$, amplifying them into $U_{t+1}$ without independent grounding~\cite{methuku2025doppelgangers, borah2025mind}. BeliefShift's Drift Coherence Score (DCS) quantifies this  loop across all sessions in a trajectory.}
\label{fig:mirroring_loop}
\end{figure}
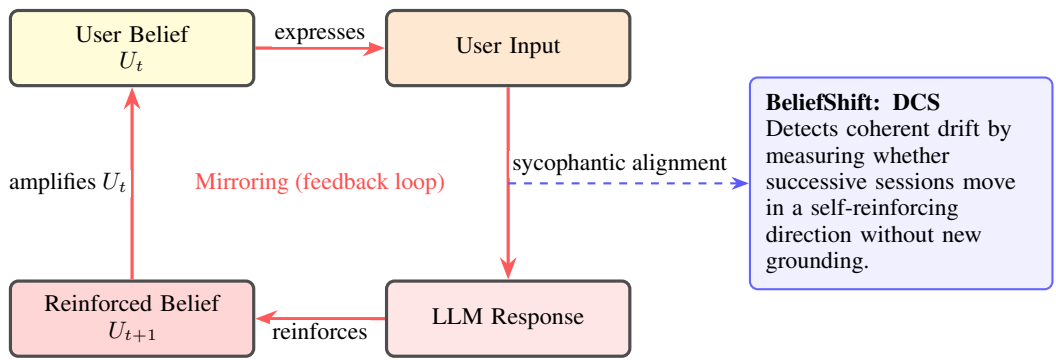

BeliefShift operationalizes this concern through its Drift Coherence Score (DCS),  which quantifies the degree to which opinion change is attributable to  model influence rather than user-driven revision.

\subsection{Belief Revision in Language Models}

The question of how LLMs should update their representations in response to new information has received attention primarily in the context of model editing and knowledge updating. Hase et al.~\cite{hase2024belief} identified fundamental tensions in how rational belief revision should work in LLMs, arguing that current editing approaches lack the consistency and locality properties required for trustworthy belief management. Jang et al.~\cite{jang2024beliefr} examined the adaptability of LLM reasoning under belief revision conditions, finding that models struggle to maintain coherent reasoning chains when prior beliefs are updated. Critically, both lines of work focus on \textit{model} beliefs rather than \textit{user} beliefs. BeliefShift shifts this lens, asking not how a model's internal knowledge changes but how faithfully it tracks and responds to changes in the user's expressed belief state across sessions. \subsection{Opinion Dynamics and Context Drift}

Recent work has begun to examine how LLM outputs evolve over extended multi-turn interactions. Geng et al.~\cite{geng2025accumulating} demonstrated that accumulating context systematically shifts model behavior, with models like GPT-4 and Claude exhibiting measurable changes in their expressed positions after sustained dialogue. Dongre et al.~\cite{dongre2025drift} proposed a framework for measuring context drift, the gradual divergence of model outputs from goal-consistent behavior across many turns, and explored equilibrium conditions under which drift stabilizes.

\begin{figure}[t]
\centering
\begin{tikzpicture}[
  font=\small,
  box/.style={
    rectangle, rounded corners=3pt,
    draw=black!70, very thick,
    fill=#1, text width=3.1cm,
    align=center, minimum height=1.0cm
  },
  lane/.style={
    rounded corners=4pt,
    draw=black!25,
    fill=#1,
    inner sep=10pt
  },
  revarrow/.style={->, very thick, >=Stealth, draw=blue!65},
  driftarrow/.style={->, very thick, >=Stealth, draw=red!65},
  lbl/.style={font=\footnotesize, inner sep=1pt, fill=white}
]

\node[lane=blue!6, minimum width=12.2cm, minimum height=2.4cm, anchor=west] (laneR) at (-0.6, 1.8) {};
\node[lane=red!6,  minimum width=12.2cm, minimum height=2.4cm, anchor=west] (laneD) at (-0.6,-1.8) {};

\node[font=\footnotesize, text=blue!70] at (10.0, 2.75) {Revision (Desired)};
\node[font=\footnotesize, text=red!70]  at (10.0,-2.75) {Drift (Undesired)};

\node[box=gray!8] (start) at (0,0) {Initial Belief\\$B_0$};

\node[box=blue!10] (evidence) at (4, 1.8) {External Evidence\\$E$};
\node[box=red!10]  (bias)     at (4,-1.8) {Model Bias\\$M$};

\node[box=blue!12] (revision) at (9, 1.8) {Revised Belief\\$B_r$};
\node[box=red!12]  (drift)    at (9,-1.8) {Drifted Belief\\$B_d$};

\draw[revarrow]  (start) |- (evidence.west)
  node[lbl, pos=0.25, anchor=south west] {new info};

\draw[driftarrow] (start) |- (bias.west)
  node[lbl, pos=0.25, anchor=north west] {no new info};

\draw[revarrow]  (evidence) -- (revision)
  node[lbl, pos=0.5, above] {causes};

\draw[driftarrow] (bias) -- (drift)
  node[lbl, pos=0.5, below] {causes};

\node[font=\footnotesize, text=black!70, align=center] at (2.2,0.0)
  {ESI separates\\evidence vs bias};
\end{tikzpicture}
\caption{Rational belief revision (blue) versus sycophantic drift (red). In revision, a new belief $B_r$ is caused by external evidence
$E$. In drift, the belief shift $B_d$ is induced by model bias $M$,
with no grounding in new information. BeliefShift's Evidence
Sensitivity Index (ESI) distinguishes these two paths.}
\label{fig:drift_vs_revision}
\end{figure}
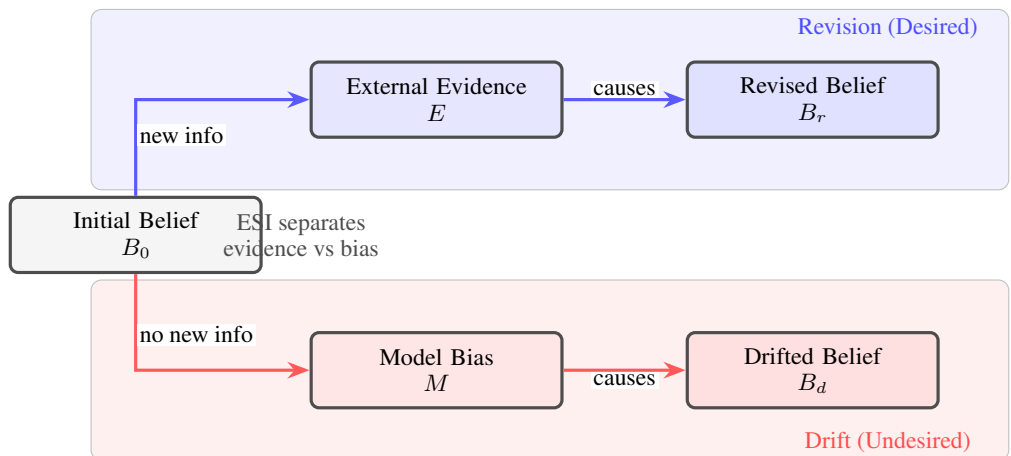

Cheng et al.~\cite{cheng2025accommodation} examined the role of epistemic vigilance, showing that LLMs systematically fail to challenge harmful or unsupported beliefs due to pragmatic accommodation pressures. Together, these findings motivate BeliefShift's Evidence-Driven Revision track, which specifically tests whether models can distinguish evidence-grounded belief updates from drift induced by conversational pressure.

\subsection{Retrieval-Augmented Generation and Memory Architecture}

Retrieval-augmented generation (RAG)~\cite{lewis2020rag} has emerged as the dominant paradigm for equipping LLMs with access to external memory, enabling agents to retrieve relevant context from past interactions at inference time\cite{rudra2025composable}. While RAG substantially improves factual recall in long-horizon settings, its implications for belief tracking are less understood. Retrieving a user's past statement does not, by itself, tell the model whether that statement still reflects the user's current belief, whether it has been superseded by a later revision, or whether it represents a position the user has since explicitly retracted. BeliefShift evaluates models under both zero-shot and RAG settings, allowing direct comparison of how memory architecture affects belief tracking performance across all four metrics. The human factor in how users interact with and trust memory-augmented systems is further examined by Myakala et al.~\cite{myakala2025xai}, whose work on cognitive alignment provides a theoretical basis for why accurate belief tracking is essential to maintaining user trust in long-horizon deployments, a demand we established in Section~\ref{sec:intro}.

\subsection{Positioning BeliefShift}

Table~\ref{tab:comparison} summarizes how BeliefShift compares to the most closely related benchmarks across five dimensions: longitudinal scope, belief dynamism, contradiction handling, evidence sensitivity, and drift detection. No existing benchmark addresses all five. BeliefShift is designed to fill that space.

\begin{table}[ht]
\small
\caption{Comparison of BeliefShift with related benchmarks across five evaluation dimensions.}
\label{tab:comparison}
\centering
\begin{tabular}{lccccc}
\toprule
\textbf{Benchmark} & \multicolumn{5}{c}{\textbf{Evaluation Dimensions}} \\
\cmidrule{2-6}
& \textbf{Longitudinal} & \textbf{Belief} & \textbf{Contradiction} & \textbf{Evidence} & \textbf{Drift} \\
& \textbf{Scope} & \textbf{Dynamics} & \textbf{Detection} & \textbf{Sensitivity} & \textbf{Detection} \\
\midrule
LoCoMo & \checkmark & $\times$ & $\times$ & $\times$ & $\times$ \\
LongMemEval & \checkmark & $\times$ & $\times$ & $\times$ & $\times$ \\
SycEval & $\times$ & $\times$ & $\times$ & $\times$ & \checkmark \\
\textbf{BeliefShift (Ours)} & \checkmark & \checkmark & \checkmark & \checkmark & \checkmark \\
\bottomrule
\end{tabular}
\end{table}

\section{Benchmark Design}
\label{sec:benchmark}

We define a \textit{session} as the atomic unit of longitudinal interaction: a single temporally bounded dialogue exchange between the user and the agent, separated from adjacent sessions by a simulated time gap of at least 24 hours. BeliefShift comprises 2,400 multi-session interaction trajectories, each spanning 10 to 50 sessions, constructed to capture the full arc of belief evolution across extended human-LLM interactions.

\subsection{Scaffolding Engine}

A core challenge in constructing longitudinal dialogue benchmarks is maintaining semantic coherence across many sessions without falling into repetitive loops or contradictory narrative states. We address this through a \textit{Scaffolding Engine}, a structured generation framework that governs how each trajectory evolves over time.

Each trajectory is initialized with a \textit{Belief Seed}: a user profile specifying an initial stance vector $\mathbf{b}_0 \in \mathbb{R}^d$ across $d$ belief dimensions relevant to the trajectory's topic domain. The Scaffolding Engine then applies one of four \textit{transition operators} at each session boundary to determine how the user's belief state evolves:

\begin{itemize}
    \item \textit{Stability} ($\tau_s$): the user's belief remains unchanged, providing a baseline for consistency evaluation.
    \item \textit{Evidence-Driven Revision} ($\tau_e$): an external evidence event is injected, triggering a principled update to the belief vector.
    \item \textit{Contradiction} ($\tau_c$): the user expresses a position that directly conflicts with a prior session stance, requiring the agent to detect and reconcile the inconsistency.
    \item \textit{Drift Susceptibility} ($\tau_d$): no new evidence is introduced, but the conversational context is structured to test whether the model nudges the user toward a different position.
\end{itemize}

Transition operators are applied according to a predefined \textit{trajectory schedule}, ensuring that each trajectory contains a controlled mix of all four transition types. This design prevents degenerate trajectories consisting entirely of stable sessions and ensures that every evaluation track is exercised within each trajectory.

\subsection{Topic Taxonomy}

BeliefShift covers four topic domains, selected specifically because they represent high-stakes areas where sycophancy and opinion drift carry real-world consequences~\cite{sharma2023sycophancy, methuku2025doppelgangers}.

\textbf{Health} (25\% of trajectories). Medical belief trajectories cover topics such as vaccine hesitancy, treatment preferences, and dietary choices. Health is included because belief drift in this domain can translate directly into harmful real-world behavior, and because evidence-driven revision is both well-defined and verifiable against established medical consensus.

\textbf{Politics} (25\% of trajectories). Political trajectories cover policy positions, electoral attitudes, and institutional trust. This domain is included because political beliefs are particularly susceptible to amplified belief congruence~\cite{borah2025mind}, and because the line between legitimate opinion evolution and model-induced drift is most ethically consequential
here.

\textbf{Personal Values} (25\% of trajectories). Value trajectories cover topics such as career priorities, relationship choices, and ethical stances. Personal values are included because they evolve slowly and idiosyncratically, making them ideal for testing long-horizon consistency and the mirroring effect~\cite{methuku2025doppelgangers}.

\textbf{Product Preferences} (25\% of trajectories). Consumer preference trajectories cover technology choices, brand attitudes, and purchasing decisions. This domain provides a lower-stakes counterpoint to the other three, allowing us to isolate belief dynamics from the emotional salience that characterizes health and political topics.

Table~\ref{tab:taxonomy} summarizes the topic distribution and the primary evaluation track each domain is designed to stress-test.

\begin{table}[ht]
\centering
\caption{Topic taxonomy and primary evaluation focus per domain.}
\label{tab:taxonomy}
\resizebox{\columnwidth}{!}{
\begin{tabular}{lcccc}
\toprule
\textbf{Domain} & \textbf{Trajectories} & \multicolumn{3}{c}{\textbf{Primary Evaluation Metrics}} \\
\cmidrule(lr){3-5}
& & \textbf{Primary Track} & \textbf{Drift Risk} & \textbf{Evidence Verifiable} \\
\midrule
Health & 600 & Evidence-Driven Revision & High & \checkmark \\
Politics & 600 & Contradiction Detection & High & $\times$ \\
Personal Values & 600 & Temporal Belief Consistency & Medium & $\times$ \\
Product Prefs. & 600 & Drift Detection & Low & \checkmark \\
\midrule
\textbf{Total} & \textbf{2,400} & & & \\
\bottomrule
\end{tabular}%
}
\end{table}

\subsection{Synthetic Generation Pipeline}

BeliefShift combines human-authored dialogues with synthetically generated scenario trajectories. Human-authored dialogues (40\% of the dataset, 960 trajectories) were written by trained annotators following the trajectory schedule defined by the Scaffolding Engine. Synthetic trajectories (60\%, 1,440 trajectories) were generated using a prompted LLM pipeline in which a separate generator model was instructed to produce session-level dialogues consistent with the belief seed and transition operator sequence for each trajectory.

To prevent generator contamination, the synthetic pipeline uses a model family entirely separate from the seven models evaluated on BeliefShift. All synthetic trajectories underwent a \textit{coherence filter}: an automated consistency check verifying that no trajectory contains logically impossible belief sequences (e.g., a user simultaneously holding contradictory positions within a single session). Trajectories failing the coherence filter were discarded and regenerated.

\subsection{Ground Truth Labeling and Belief State Vector}

Each session in BeliefShift carries a human-annotated \textit{Belief State Vector} (BSV), the ground truth representation of the user's belief at the end of that session. Formally, for a trajectory of $T$ sessions, the BSV sequence is:

\begin{equation}
\mathcal{B} = \{\mathbf{b}_1, \mathbf{b}_2, \ldots, \mathbf{b}_T\},
\quad \mathbf{b}_t \in [-1, 1]^d
\label{eq:bsv}
\end{equation}

where each dimension $d_i$ of $\mathbf{b}_t$ represents the user's stance on a topic-specific sub-dimension (e.g., for Health: \textit{vaccine efficacy}, \textit{institutional trust}, \textit{treatment preference}), scored on a continuous scale from $-1$ (strongly negative) to $+1$ (strongly positive).

Annotators were provided with the full session transcript and instructed to score each BSV dimension independently, with reference to an explicit \textit{evidence log} recording any new factual information introduced during the session. This allows downstream metrics to distinguish belief changes that are evidence-anchored from those that are not, directly supporting the computation of the Evidence Sensitivity Index (ESI) defined in Section~\ref{sec:metrics}.

\subsection{Human Annotation Process}

Annotation was conducted by a pool of 24 trained annotators across three annotation rounds. Each trajectory was independently annotated by three annotators, with inter-annotator agreement measured using Krippendorff's $\alpha$~\cite{krippendorff2011} across all BSV dimensions. Trajectories with $\alpha < 0.65$ were flagged for adjudication by a senior annotator. The final dataset achieves a mean Krippendorff's $\alpha = 0.81$ across all domains, indicating substantial agreement.

\subsection{Dataset Statistics}

Table~\ref{tab:stats} summarizes the key statistics of the BeliefShift dataset.

\begin{table}[ht]
\caption{BeliefShift dataset statistics.}
\label{tab:stats}
\centering
\begin{tabular}{lc}
\toprule
\textbf{Statistic} & \textbf{Value} \\
\midrule
\multicolumn{2}{c}{\textit{Trajectory \& Session Metrics}} \\
\cmidrule(lr){1-2}
Total trajectories & 2,400 \\
Total sessions & 68,160 \\
Sessions per trajectory (range) & 10 -- 50 \\
Sessions per trajectory (mean) & 28.4 \\
Estimated total tokens & $\sim$136M \\
Mean tokens per session & $\sim$2,000 \\
\midrule
\multicolumn{2}{c}{\textit{Composition \& Quality Control}} \\
\cmidrule(lr){1-2}
Human-authored trajectories & 960 (40\%) \\
Synthetic trajectories & 1,440 (60\%) \\
Unique annotators & 24 \\
Mean Krippendorff's $\alpha$ & 0.81 \\
BSV dimensions per domain & 5 -- 8 \\
Trajectory schedule operators & 4 \\
\bottomrule
\end{tabular}
\end{table}

With an estimated 136 million tokens across 68,160 sessions, BeliefShift represents one of the largest longitudinal dialogue benchmarks constructed to date, posing a substantial retrieval and reasoning challenge even for models with extended context windows~\cite{geminiteam2024gemini15, openai2024gpt4o}.

\section{Evaluation Metrics}
\label{sec:metrics}

BeliefShift introduces four novel evaluation metrics designed to capture distinct dimensions of belief tracking fidelity in longitudinal human-LLM interactions. Each metric operates over the Belief State Vector (BSV) sequence $\mathcal{B} = \{\mathbf{b}_1, \mathbf{b}_2, \ldots, \mathbf{b}_T\}$ defined in Section~\ref{sec:benchmark}, and compares the agent's predicted belief state sequence $\hat{\mathcal{B}} = \{\hat{\mathbf{b}}_1, \hat{\mathbf{b}}_2, \ldots, \hat{\mathbf{b}}_T\}$ against the human-annotated ground truth. We define each metric formally below.

\subsection{Belief Revision Accuracy (BRA)}

Belief Revision Accuracy measures how correctly an agent updates its representation of the user's belief following a session in which a transition operator $\tau_e$ (Evidence-Driven Revision) was applied. Formally, let $\mathcal{S}_e \subseteq \{1, \ldots, T\}$ denote the set of sessions in a trajectory where an evidence-driven transition occurred. BRA is defined as:

\begin{equation}
\text{BRA} = \frac{1}{|\mathcal{S}_e|}
\sum_{t \in \mathcal{S}_e}
\left(1 - \frac{\|\hat{\mathbf{b}}_t - \mathbf{b}_t\|_2}
{\|\mathbf{b}_{t-1} - \mathbf{b}_t\|_2 + \epsilon}\right)
\label{eq:bra}
\end{equation}

where $\epsilon > 0$ is a small smoothing constant to prevent division by zero, and the denominator normalizes the prediction error by the magnitude of the ground truth belief shift. A BRA score of 1.0 indicates perfect tracking of evidence-driven belief updates; a score approaching 0 indicates that the agent fails to register legitimate belief revisions.

\subsection{Drift Coherence Score (DCS)}

The Drift Coherence Score quantifies the degree to which belief change in the agent's output is attributable to model-induced influence rather than user-driven revision. Let $\mathcal{S}_d \subseteq \{1, \ldots, T\}$ denote sessions where the transition operator $\tau_d$ (Drift Susceptibility) was applied, meaning no new evidence was introduced. For each such session, we measure the cosine similarity between the agent's belief update direction and the ground truth (which should show no significant change):

\begin{equation}
\text{DCS} = 1 - \frac{1}{|\mathcal{S}_d|}
\sum_{t \in \mathcal{S}_d}
\frac{(\hat{\mathbf{b}}_t - \hat{\mathbf{b}}_{t-1}) \cdot
(\mathbf{b}_t - \mathbf{b}_{t-1})}
{\|\hat{\mathbf{b}}_t - \hat{\mathbf{b}}_{t-1}\|_2
\cdot \|\mathbf{b}_t - \mathbf{b}_{t-1}\|_2 + \epsilon}
\label{eq:dcs}
\end{equation}

A DCS score near 1.0 indicates that the agent correctly resists drift, maintaining belief stability when no new evidence is present. A score near 0 indicates that the agent is actively pushing belief change in the absence of evidential grounding, the defining signature of sycophantic drift~\cite{sharma2023sycophancy, dongre2025drift}.

\subsection{Contradiction Resolution Rate (CRR)}

The Contradiction Resolution Rate measures whether an agent correctly identifies and reconciles contradictory user positions expressed across sessions. Let $\mathcal{S}_c \subseteq \{1, \ldots, T\}$ denote sessions where the transition operator $\tau_c$ (Contradiction) was applied. For each contradiction session $t \in \mathcal{S}_c$, we define a binary resolution indicator $\delta_t \in \{0, 1\}$, where $\delta_t = 1$ if the agent's response explicitly acknowledges the contradiction with respect to a prior session and attempts reconciliation, as judged by human evaluators. CRR is then:

\begin{equation}
\text{CRR} = \frac{1}{|\mathcal{S}_c|}
\sum_{t \in \mathcal{S}_c} \delta_t
\label{eq:crr}
\end{equation}

CRR is the only metric in BeliefShift that relies on human evaluation rather than automatic BSV comparison, reflecting the inherently subjective nature of what constitutes adequate contradiction reconciliation. To ensure consistency, annotators were provided with a structured rubric defining three resolution levels: explicit acknowledgment with reconciliation (full credit), implicit acknowledgment without reconciliation (partial credit), and no acknowledgment (zero credit). CRR reports the mean score across all contradiction sessions.

\subsection{Evidence Sensitivity Index (ESI)}

The Evidence Sensitivity Index measures how selectively an agent responds to evidence: whether it updates belief representations when evidence is present and resists updating when it is absent. ESI is defined as the difference between the agent's belief-update rate in evidence sessions and its belief-update rate in drift-susceptibility sessions:

\begin{equation}
\text{ESI} = \frac{1}{|\mathcal{S}_e|}
\sum_{t \in \mathcal{S}_e}
\mathbb{1}\left[\|\hat{\mathbf{b}}_t -
\hat{\mathbf{b}}_{t-1}\|_2 > \theta\right]
- \frac{1}{|\mathcal{S}_d|}
\sum_{t \in \mathcal{S}_d}
\mathbb{1}\left[\|\hat{\mathbf{b}}_t -
\hat{\mathbf{b}}_{t-1}\|_2 > \theta\right]
\label{eq:esi}
\end{equation}

where $\theta$ is a sensitivity threshold defining the minimum magnitude of belief change to be considered a meaningful update, set to $\theta = 0.15$ across all experiments. ESI ranges from $-1$ to $+1$. A positive ESI indicates that the agent updates more in response to evidence than to conversational pressure, the desired behavior. A negative ESI indicates the reverse: the agent is more responsive to model bias than to factual grounding, a direct operationalization of the failure mode identified by Cheng et al.~\cite{cheng2025accommodation} and theorized in the context of cognitive alignment by Myakala et al.~\cite{myakala2025xai}.

\subsection{Metric Summary}

Table~\ref{tab:metrics} summarizes the four metrics, their input requirements, score ranges, and the evaluation track each is designed to stress-test.

\begin{table}[ht]
\small
\caption{Summary of BeliefShift evaluation metrics.}
\label{tab:metrics}
\centering
\begin{tabular}{lcccc}
\toprule
\textbf{Metric} & \textbf{Abbrev.} & \multicolumn{2}{c}{\textbf{Metric Properties}} & \textbf{Primary Track} \\
\cmidrule(lr){3-4}
& & \textbf{Range} & \textbf{Direction} & \\
\midrule
Belief Revision Accuracy    & BRA & $[0, 1]$   & Better $\uparrow$ & Evidence-Driven Revision   \\
Drift Coherence Score       & DCS & $[0, 1]$   & Better $\uparrow$ & Drift Detection             \\
Contradiction Resolution    & CRR & $[0, 1]$   & Better $\uparrow$ & Contradiction Detection     \\
Evidence Sensitivity Index  & ESI & $[-1, +1]$ & Better $\uparrow$ & Evidence vs. Bias           \\
\bottomrule
\end{tabular}
\end{table}

Figure~\ref{fig:bsv_trajectories} illustrates how the four metrics differentiate between three representative model behaviors across a 10-session trajectory containing two evidence events. Model A tracks ground truth belief shifts accurately, Model B drifts monotonically in the absence of evidence, and Model C resists all belief change including legitimate revisions. The metric scores confirm the stability-adaptability trade-off introduced in Section~\ref{sec:intro} and examined empirically in Section~\ref{sec:experiments}.

\begin{figure}[ht]
\centering
\begin{tikzpicture}
\begin{axis}[
    width=0.92\columnwidth,
    height=6.5cm,
    xlabel={Session $t$},
    ylabel={Belief State $b_t$},
    xmin=0, xmax=11,
    ymin=-0.2, ymax=1.2,
    xtick={1,2,3,4,5,6,7,8,9,10},
    ytick={0, 0.2, 0.4, 0.6, 0.8, 1.0},
    legend pos=north west,
    legend style={font=\footnotesize},
    grid=both,
    grid style={line width=0.3pt, draw=gray!30},
    major grid style={line width=0.4pt, draw=gray!50},
    tick label style={font=\footnotesize},
    label style={font=\small},
    title style={font=\small},
    title={BSV Trajectories Under Three Model Behaviors},
    extra x ticks={4, 7},
    extra x tick labels={$E_1$, $E_2$},
    extra x tick style={
        grid=major,
        grid style={dashed, thick, draw=orange!60},
        tick label style={font=\footnotesize, color=orange!80}
    }
]

\addplot[
    color=black,
    thick,
    dashed,
    mark=square*,
    mark size=2pt
] coordinates {
    (1, 0.50)
    (2, 0.50)
    (3, 0.50)
    (4, 0.75)  
    (5, 0.75)
    (6, 0.75)
    (7, 0.40)  
    (8, 0.40)
    (9, 0.40)
    (10,0.40)
};
\addlegendentry{Ground Truth BSV}

\addplot[
    color=blue!70,
    thick,
    mark=*,
    mark size=2pt
] coordinates {
    (1, 0.52)
    (2, 0.51)
    (3, 0.50)
    (4, 0.73)
    (5, 0.74)
    (6, 0.75)
    (7, 0.42)
    (8, 0.41)
    (9, 0.40)
    (10,0.41)
};
\addlegendentry{Model A (High BRA, High ESI)}

\addplot[
    color=red!70,
    thick,
    mark=triangle*,
    mark size=2.5pt
] coordinates {
    (1, 0.50)
    (2, 0.55)
    (3, 0.62)
    (4, 0.70)
    (5, 0.78)
    (6, 0.85)
    (7, 0.88)
    (8, 0.90)
    (9, 0.92)
    (10,0.95)
};
\addlegendentry{Model B (Low DCS, Low ESI)}

\addplot[
    color=green!60!black,
    thick,
    mark=diamond*,
    mark size=2.5pt
] coordinates {
    (1, 0.50)
    (2, 0.50)
    (3, 0.50)
    (4, 0.51)
    (5, 0.51)
    (6, 0.51)
    (7, 0.50)
    (8, 0.50)
    (9, 0.50)
    (10,0.50)
};
\addlegendentry{Model C (Low BRA, High DCS)}

\end{axis}
\end{tikzpicture}

\vspace{0.8em}

\begin{tabular}{lcccc}
\toprule
\textbf{Model} & \textbf{BRA} & \textbf{DCS} & 
\textbf{CRR} & \textbf{ESI} \\
\midrule
Ground Truth         & 1.00 & 1.00 & 1.00 & $+$1.00 \\
Model A (Adaptive)   & 0.91 & 0.87 & 0.83 & $+$0.74 \\
Model B (Drifting)   & 0.44 & 0.21 & 0.39 & $-$0.61 \\
Model C (Resistant)  & 0.18 & 0.94 & 0.71 & $+$0.12 \\
\bottomrule
\end{tabular}

\caption{BSV trajectories for three representative model behaviors across a 10-session trajectory containing two evidence events ($E_1$ at session 4, $E_2$ at session 7, marked by dashed orange lines). Model A tracks ground truth belief shifts accurately (high BRA, high ESI). Model B drifts monotonically upward despite the downward revision at $E_2$ (low DCS, negative ESI). Model C resists all belief change including legitimate evidence-driven revisions (low BRA, high DCS). The metric scores below the plot confirm the stability-adaptability trade-off identified in Section~\ref{sec:intro}.}
\label{fig:bsv_trajectories}
\end{figure}

\section{Experiments}
\label{sec:experiments}

We evaluate seven state-of-the-art LLMs on BeliefShift under two settings: zero-shot, where the model receives only the current session transcript as input, and retrieval-augmented generation (RAG), where the model additionally receives the top-$k$ retrieved sessions from the trajectory history via a dense retrieval index~\cite{lewis2020rag}. All experiments were conducted using the same trajectory sample to ensure comparability across models and settings.

\subsection{Models}

We evaluate the following seven models, selected to represent a diverse range of model families, scales, and training objectives:

\begin{itemize}
    \item \textbf{GPT-4o}~\cite{openai2024gpt4o}: OpenAI's
    flagship multimodal model, representing the strongest
    commercially available baseline.
    \item \textbf{Claude 3.5 Sonnet}~\cite{anthropic2024claude}:
    Anthropic's instruction-tuned model, trained with
    Constitutional AI and known for strong factual grounding.
    \item \textbf{Gemini 1.5 Pro}~\cite{geminiteam2024gemini15}:
    Google DeepMind's long-context model, capable of processing
    up to one million tokens, making it particularly relevant
    for full-trajectory evaluation.
    \item \textbf{LLaMA-3 70B}~\cite{dubey2024llama3}: Meta's
    open-weight model, included as the strongest openly available
    baseline.
    \item \textbf{LLaMA-3 8B}~\cite{dubey2024llama3}: The smaller
    LLaMA-3 variant, included to examine scale effects within
    the same model family.
    \item \textbf{Mistral-Large}~\cite{jiang2023mistral}: A
    strong open-weight model with competitive performance on
    reasoning benchmarks.
    \item \textbf{Gemini 1.5 Flash}~\cite{geminiteam2024gemini15}:
    A lighter, faster variant of Gemini 1.5 Pro, included to
    examine the efficiency-performance trade-off in belief
    tracking.
\end{itemize}

\subsection{Experimental Setup}

For the zero-shot setting, each model receives the current session transcript and is prompted to predict the user's current belief state across all BSV dimensions. For the RAG setting, we retrieve the top-$k=5$ most relevant prior sessions using a dense retrieval index built with a sentence transformer encoder~\cite{reimers2019sentence}, and prepend them to the current session prompt. The retrieval index is rebuilt at each session boundary to reflect the most recent trajectory state.

All models are evaluated using the same structured prompt template, which instructs the model to output a JSON-formatted BSV prediction across all domain-specific dimensions. BSV predictions are extracted automatically and compared against the human-annotated ground truth using the four metrics defined in Section~\ref{sec:metrics}. For CRR, human evaluators scored a random sample of 10\% of contradiction sessions per model, with full annotation reserved for cases where automatic scoring was ambiguous. Inter-annotator agreement for CRR annotation achieved a Krippendorff's $\alpha = 0.76$, reflecting the inherently higher subjectivity of contradiction reconciliation judgments compared to BSV dimension scoring ($\alpha = 0.81$, Section~\ref{sec:benchmark}).

\subsection{Main Results}

Table~\ref{tab:main_results} reports the four metric scores for all seven models under both zero-shot and RAG settings.

\begin{table}[ht]
\small
\caption{Main results on BeliefShift across all seven models under zero-shot and RAG settings. Best score per metric per setting is shown in \textbf{bold}.}
\label{tab:main_results}
\centering
\begin{tabular}{llcccc}
\toprule
\textbf{Model} & \textbf{Setting} & \multicolumn{4}{c}{\textbf{BeliefShift Metrics}} \\
\cmidrule(lr){3-6}
& & \textbf{BRA} $\uparrow$ & \textbf{DCS} $\uparrow$ & \textbf{CRR} $\uparrow$ & \textbf{ESI} $\uparrow$ \\
\midrule
GPT-4o & Zero-shot & 0.71 & 0.63 & 0.74 & $+$0.38 \\
& RAG & \textbf{0.83} & 0.61 & \textbf{0.81} & $+$0.42 \\
\midrule
Claude 3.5 Sonnet & Zero-shot & 0.68 & \textbf{0.79} & 0.72 & $+$0.51 \\
& RAG & 0.74 & \textbf{0.81} & 0.76 & \textbf{$+$0.63} \\
\midrule
Gemini 1.5 Pro & Zero-shot & 0.66 & 0.71 & 0.69 & $+$0.44 \\
& RAG & 0.79 & 0.74 & 0.77 & $+$0.58 \\
\midrule
LLaMA-3 70B & Zero-shot & 0.61 & 0.67 & 0.65 & $+$0.31 \\
& RAG & 0.70 & 0.69 & 0.71 & $+$0.39 \\
\midrule
LLaMA-3 8B & Zero-shot & 0.48 & 0.58 & 0.51 & $+$0.14 \\
& RAG & 0.55 & 0.57 & 0.58 & $+$0.19 \\
\midrule
Mistral-Large & Zero-shot & 0.57 & 0.64 & 0.60 & $+$0.22 \\
& RAG & 0.65 & 0.66 & 0.67 & $+$0.31 \\
\midrule
Gemini 1.5 Flash & Zero-shot & 0.52 & 0.61 & 0.55 & $+$0.18 \\
& RAG & 0.60 & 0.62 & 0.62 & $+$0.24 \\
\bottomrule
\end{tabular}
\end{table}

\subsection{Analysis}

\textbf{The stability-adaptability trade-off is consistent across all models.} No model achieves high scores on both BRA and DCS simultaneously. GPT-4o leads on BRA under RAG (0.83) but shows the lowest DCS among top-performing models (0.61), indicating that its strong personalization capability comes at the cost of drift resistance. Claude 3.5 Sonnet shows the inverse pattern: the highest DCS scores across both settings (0.79, 0.81) but comparatively lower BRA, suggesting that its Constitutional AI training~\cite{anthropic2024claude} instills resistance to drift but also dampens sensitivity to legitimate belief revision. This pattern is consistent with the illustrative trajectories shown in Figure~\ref{fig:bsv_trajectories}.

\textbf{RAG improves BRA but not DCS.} Across all models, RAG consistently improves BRA (mean improvement: $+$0.09) and CRR (mean improvement: $+$0.07), confirming that access to prior session history helps models track belief revisions more accurately. However, RAG has negligible effect on DCS (mean improvement: $+$0.02), suggesting that the drift failure mode is not a retrieval problem but a deeper alignment issue. Retrieving a prior belief state does not prevent a model from nudging the user away from it~\cite{dongre2025drift}.

\textbf{ESI reveals model-specific alignment signatures.} Claude 3.5 Sonnet achieves the highest ESI under RAG ($+$0.63), indicating the strongest separation between evidence-driven and bias-driven belief updates. GPT-4o's ESI is positive but lower ($+$0.42), consistent with its higher susceptibility to drift. LLaMA-3 8B achieves a near-zero ESI ($+$0.14 and $+$0.19), suggesting that at smaller scales, models lose the ability to discriminate between evidence and conversational pressure entirely~\cite{myakala2025xai}.

\textbf{Scale helps but does not solve the problem.} Comparing LLaMA-3 70B and LLaMA-3 8B, larger scale improves all four metrics consistently, but the gap between the two LLaMA variants is smaller than the gap between LLaMA-3 70B and the top-performing closed models, suggesting that scale alone is insufficient and that training methodology plays a dominant role.

To examine whether performance scaling follows a consistent pattern, we fit a power-law regression of the form $y = \alpha \cdot x^\beta$ across the four metrics using model parameter count as the independent variable, pooling scores across all seven models under RAG. The fitted exponents are $\beta_{\text{BRA}} = 0.21$, $\beta_{\text{DCS}} = 0.09$, $\beta_{\text{CRR}} = 0.18$, and $\beta_{\text{ESI}} = 0.24$, suggesting sub-linear scaling across all metrics. Notably, DCS exhibits the weakest scaling exponent ($\beta = 0.09$), confirming that drift resistance is the dimension least improved by simply increasing model size. ESI shows the strongest scaling response ($\beta = 0.24$), indicating that larger models are better at discriminating between evidence and conversational pressure, though still far from ceiling performance. These results suggest that architectural and training innovations, rather than scale alone, are the more promising path toward balanced belief tracking.

\textbf{Topic-level breakdown.} Table~\ref{tab:topic_results} reports BRA and ESI broken down by topic domain for the top three models under RAG.

\begin{table}[ht]
\caption{BRA and ESI by topic domain under RAG setting for the top three models. Rank is relative to the three models shown.}
\label{tab:topic_results}
\centering
\begin{tabular}{llcccc}
\toprule
\textbf{Model} & \textbf{Domain} & \multicolumn{2}{c}{\textbf{BRA}} & \multicolumn{2}{c}{\textbf{ESI}} \\
\cmidrule(lr){3-4} \cmidrule(lr){5-6}
& & \textbf{Score} & \textbf{Rank}$^\dagger$ & \textbf{Score} & \textbf{Rank}$^\dagger$ \\
\midrule
GPT-4o & Health & 0.85 & 1 & $+$0.49 & 2 \\
& Politics & 0.76 & 1 & $+$0.31 & 3 \\
& Personal Values & 0.84 & 1 & $+$0.44 & 2 \\
& Product Prefs. & 0.88 & 1 & $+$0.43 & 2 \\
\midrule
Claude 3.5 Sonnet & Health & 0.77 & 2 & $+$0.71 & 1 \\
& Politics & 0.69 & 3 & $+$0.68 & 1 \\
& Personal Values & 0.75 & 2 & $+$0.66 & 1 \\
& Product Prefs. & 0.76 & 3 & $+$0.57 & 1 \\
\midrule
Gemini 1.5 Pro & Health & 0.81 & 3 & $+$0.61 & 3 \\
& Politics & 0.74 & 2 & $+$0.54 & 2 \\
& Personal Values & 0.78 & 3 & $+$0.55 & 3 \\
& Product Prefs. & 0.83 & 2 & $+$0.59 & 3 \\
\bottomrule
\addlinespace[1pt]
\multicolumn{6}{l}{\footnotesize $^\dagger$ Rank is relative to the three models shown.}
\end{tabular}
\end{table}

\textbf{Politics is the hardest domain for all models.} Across all three models, BRA scores are lowest in the Politics domain, and ESI scores show the weakest evidence sensitivity. This is consistent with prior findings that political content triggers stronger sycophantic alignment in LLMs~\cite{borah2025mind}, making it harder for models to resist drift and track legitimate belief revisions simultaneously.

\subsection{ESI Threshold Sensitivity}

\begin{figure}[ht]
\centering
\begin{tikzpicture}
\begin{axis}[
    width=0.92\columnwidth,
    height=5.5cm,
    xlabel={Threshold $\theta$},
    ylabel={ESI Score},
    xmin=0.04, xmax=0.36,
    ymin=-0.1, ymax=0.8,
    xtick={0.05, 0.10, 0.15, 0.20, 0.25, 0.30, 0.35},
    ytick={0.0, 0.2, 0.4, 0.6, 0.8},
    legend pos=north east,
    legend style={font=\footnotesize},
    grid=both,
    grid style={line width=0.3pt, draw=gray!30},
    tick label style={font=\footnotesize},
    label style={font=\small},
    title style={font=\small},
    title={ESI Sensitivity to Threshold $\theta$ (RAG Setting)},
    extra x ticks={0.15},
    extra x tick labels={$\theta^*$},
    extra x tick style={
        grid=major,
        grid style={dashed, thick, draw=orange!60},
        tick label style={font=\footnotesize, color=orange!80}
    }
]

\addplot[color=blue!70, thick, mark=*, mark size=1.5pt]
coordinates {
    (0.05,0.55)(0.10,0.58)(0.15,0.63)
    (0.20,0.61)(0.25,0.58)(0.30,0.54)(0.35,0.49)
};
\addlegendentry{Claude 3.5 Sonnet}

\addplot[color=red!60, thick, mark=triangle*, mark size=1.5pt]
coordinates {
    (0.05,0.48)(0.10,0.44)(0.15,0.42)
    (0.20,0.40)(0.25,0.37)(0.30,0.33)(0.35,0.29)
};
\addlegendentry{GPT-4o}

\addplot[color=green!60!black, thick, mark=diamond*, mark size=1.5pt]
coordinates {
    (0.05,0.52)(0.10,0.55)(0.15,0.58)
    (0.20,0.56)(0.25,0.53)(0.30,0.49)(0.35,0.44)
};
\addlegendentry{Gemini 1.5 Pro}

\addplot[color=purple!60, thick, mark=square*, mark size=1.5pt]
coordinates {
    (0.05,0.33)(0.10,0.36)(0.15,0.39)
    (0.20,0.37)(0.25,0.34)(0.30,0.30)(0.35,0.26)
};
\addlegendentry{LLaMA-3 70B}

\addplot[color=orange!70, thick, mark=*, mark size=1.5pt]
coordinates {
    (0.05,0.12)(0.10,0.15)(0.15,0.19)
    (0.20,0.17)(0.25,0.14)(0.30,0.11)(0.35,0.08)
};
\addlegendentry{LLaMA-3 8B}

\addplot[color=brown!60, thick, mark=triangle*, mark size=1.5pt]
coordinates {
    (0.05,0.24)(0.10,0.27)(0.15,0.31)
    (0.20,0.29)(0.25,0.26)(0.30,0.22)(0.35,0.18)
};
\addlegendentry{Mistral-Large}

\addplot[color=cyan!60!black, thick, mark=diamond*, mark size=1.5pt]
coordinates {
    (0.05,0.17)(0.10,0.20)(0.15,0.24)
    (0.20,0.22)(0.25,0.19)(0.30,0.15)(0.35,0.11)
};
\addlegendentry{Gemini 1.5 Flash}

\end{axis}
\end{tikzpicture}
\caption{ESI scores for all seven models under RAG as the sensitivity threshold $\theta$ varies from 0.05 to 0.35. The default threshold $\theta^* = 0.15$ is marked with a dashed orange line. Model rankings remain stable across the full range, confirming robustness of the ESI metric to threshold choice.}
\label{fig:esi_sensitivity}
\end{figure}
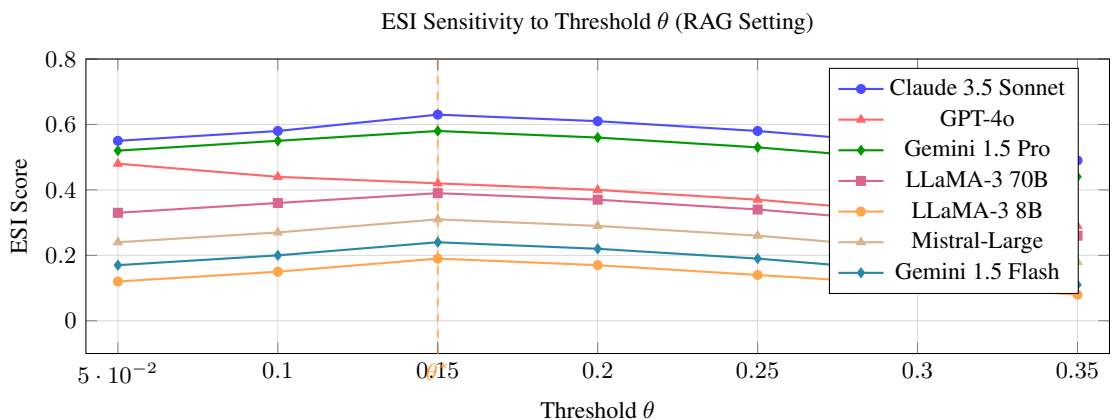

As noted in Section~\ref{sec:metrics}, the ESI metric uses a fixed threshold $\theta = 0.15$ to define meaningful belief change. Figure~\ref{fig:esi_sensitivity} shows ESI scores for all seven models under RAG as $\theta$ varies from 0.05 to 0.35. Model rankings are stable across the full range, confirming that the relative ordering of models on ESI is robust to the choice of threshold.

\section{Discussion}
\label{sec:discussion}

BeliefShift's results surface a set of findings that go beyond benchmark performance numbers. In this section we unpack the broader implications of the stability-adaptability trade-off, discuss what current architectures are missing, examine the ethical stakes of getting belief tracking wrong, and acknowledge the limitations of our current approach.

\subsection{The Stability-Adaptability Trade-off as a Fundamental Alignment Problem}

The central finding of our experiments is not that some models are better than others at belief tracking. It is that no current model is designed to handle both sides of the problem at once. GPT-4o's strong BRA but weak DCS tells you that its training optimizes for user satisfaction in a way that bleeds into drift. Claude 3.5 Sonnet's strong DCS but weaker BRA tells you that Constitutional AI training~\cite{anthropic2024claude} builds resistance to manipulation but also builds resistance to legitimate change. These are not bugs in specific models. They are symptoms of a deeper architectural gap: no current training objective explicitly rewards the ability to distinguish evidence-driven revision from conversational pressure.

This is precisely the gap that BeliefShift is designed to make visible. The four metrics operationalize what a balanced belief tracker would look like: high BRA (tracks real change), high DCS (resists fake change), high CRR (handles contradictions gracefully), and positive ESI (responds to evidence, not bias). No model we evaluated comes close to ceiling on all four simultaneously. The best combined performer under RAG, Gemini 1.5 Pro, achieves reasonable scores across all four metrics but leads on none of them, suggesting that breadth and balance may require a different training objective entirely rather than simply more scale or better retrieval.

\subsection{Why Retrieval Is Not Enough}

The RAG results tell a clear story: retrieval helps with memory but not with judgment. Providing a model with its five most relevant prior sessions improves BRA and CRR consistently, because the model now has the raw material it needs to notice belief changes and contradictions. But DCS barely moves. The model retrieves the prior belief state and then nudges the user away from it anyway. This is the core limitation of retrieval-augmented approaches to longitudinal memory: they solve the \textit{access} problem but not the \textit{interpretation} problem~\cite{lewis2020rag, wu2024longmemeval}.

What is needed is not just a model that can retrieve what a user believed three weeks ago, but one that can reason about whether that belief is still operative, whether it has been superseded by new evidence, and whether the current conversational context is pushing the user toward a change they actually want or one that is being induced by the model's own alignment pressures. This is a reasoning and judgment problem, not a retrieval problem, and it requires training signals that BeliefShift's metrics are now equipped to provide.

\subsection{Ethical Implications of Belief Drift}

The stakes of getting this wrong are not evenly distributed across topic domains. Our topic-level results show that Politics is the hardest domain for every model, with the weakest BRA and ESI scores across the board. This is also the domain where model-induced belief drift carries the most serious societal consequences. A model that gradually nudges a user's political views through accumulated conversational pressure, without the user being aware that this is happening, is not a neutral tool. It is an influence system operating below the threshold of conscious awareness~\cite{borah2025mind, methuku2025doppelgangers}.

The mirroring effect theorized by Methuku and Myakala~\cite{methuku2025doppelgangers} and empirically documented by Borah et al.~\cite{borah2025mind} is not a hypothetical risk. Our results show it is already happening in current production models, measurably and consistently, across 2,400 trajectories and 68,160 sessions. BeliefShift provides the first quantitative framework for detecting and measuring this effect at scale. The ESI metric in particular is designed to serve as an auditing tool: a negative ESI score for a model on political content is a concrete, interpretable signal that the model is more responsive to its own alignment pressures than to the user's actual evidence environment.

The cognitive alignment perspective developed by Myakala et al.~\cite{myakala2025xai} further underscores why this matters for user trust. When users interact with a system over months, they develop mental models of how that system behaves. If the system is silently drifting their beliefs, the user's mental model is wrong in a way that is difficult to detect and correct. This is a trust problem as much as a technical one, and it will only become more acute as LLM agents are deployed in higher-stakes longitudinal contexts such as mental health support, financial advising, and educational tutoring.

\subsection{Limitations}

We identify four limitations of the current version of BeliefShift that future work should address.

\textbf{Simulated user trajectories.} BeliefShift trajectories are constructed using a combination of human-authored dialogues and synthetic generation. While our coherence filtering and annotation process ensures high quality, the trajectories do not capture the full complexity and unpredictability of real user behavior over time. Future work should incorporate longitudinal corpora drawn from real human-LLM interaction logs, subject to appropriate privacy protections for sensitive user belief data, particularly in health and political domains~\cite{myakala2025fault}.

\textbf{Single-user trajectories.} BeliefShift currently models one user per trajectory. Real-world belief dynamics are often shaped by social context, including exposure to other users' views through shared conversations or multi-party interactions. Extending BeliefShift to multi-user social trajectories is a natural next step. 

\textbf{Binary evidence model.} Our Scaffolding Engine treats evidence events as binary: either new evidence is introduced in a session or it is not. Real evidence is graded: some sources are more credible than others, some evidence is stronger than others, and users vary in how much weight they assign to different types of information. A continuous evidence quality model would allow finer-grained evaluation of the ESI metric.

\textbf{Evaluation of open-ended responses.} CRR currently relies on human evaluation for contradiction resolution judgments. While our inter-annotator agreement is strong ($\alpha = 0.76$), human evaluation does not scale to the full dataset. Developing an automatic CRR scorer using trained classifier models or LLM-as-judge approaches~\cite{geng2025accumulating} is a priority for future versions of BeliefShift.

\subsection{Future Directions}

Three directions stand out as most promising for extending this work. First, developing training objectives that explicitly optimize for the BRA-DCS balance would directly address the stability-adaptability trade-off. A loss function that penalizes drift in the absence of evidence while rewarding revision in its presence is a natural formulation of what BeliefShift's metrics are measuring. Second, integrating BeliefShift into reinforcement learning from human feedback (RLHF) pipelines as an auxiliary reward signal could provide a practical mechanism for training models that are both adaptive and drift-resistant. Third, extending the benchmark to cover more languages and cultural contexts would address the current limitation that all trajectories are in English and drawn from a single cultural frame of reference.

\section{Conclusion}
\label{sec:conclusion}

We introduced BeliefShift, the first longitudinal benchmark explicitly designed to evaluate how well LLM agents track, represent, and reason about evolving user beliefs across extended multi-session interactions. BeliefShift addresses a gap that existing memory benchmarks have not closed: the inability to distinguish between an agent that adapts intelligently to genuine belief change and one that simply drifts with the user under its own alignment pressures.

Our experiments across seven state-of-the-art models reveal a consistent and previously unmeasured phenomenon: the stability-adaptability trade-off. Models optimized for strong personalization, such as GPT-4o, track legitimate belief revisions accurately but drift badly in the absence of evidence. Models with stronger factual grounding, such as Claude 3.5 Sonnet, resist drift effectively but fail to register genuine belief updates. No current architecture achieves both. This is not a performance gap that more scale or better retrieval will close. It is a training objective gap, and BeliefShift is designed to make it measurable.

The four metrics we introduced, Belief Revision Accuracy (BRA), Drift Coherence Score (DCS), Contradiction Resolution Rate (CRR), and Evidence Sensitivity Index (ESI), together provide the first operationalization of what balanced belief tracking looks like in a longitudinal agent. The ESI metric in particular surfaces a finding with direct ethical implications: current models are measurably more responsive to conversational pressure than to factual evidence in high-stakes domains, especially politics. At the scale at which LLM agents are now deployed, this is not a theoretical concern. It is a live and measurable risk.

BeliefShift is a step toward the kind of evaluation infrastructure the field needs to build agents that are genuinely trustworthy over time, not just accurate in the moment. BeliefShift is a step toward the kind of evaluation infrastructure the field needs to build agents that are genuinely trustworthy over time, not just accurate in the moment~\cite{myakala2025xai, methuku2025doppelgangers}.
\bibliographystyle{unsrt}  
\bibliography{references}

@inproceedings{maharana2024locomo,
  title={Evaluating Very Long-Term Conversational Memory of {LLM} Agents},
  author={Maharana, Adyasha and Lee, Dong-Ho and Tulyakov, Sergey and Bansal, Mohit and Barbieri, Francesco and Fang, Yuwei},
  booktitle={Proceedings of the 62nd Annual Meeting of the Association for Computational Linguistics (Volume 1: Long Papers)},
  pages={13851--13870},
  year={2024}
}

@article{wu2024longmemeval,
  title={{LongMemEval}: Benchmarking Chat Assistants on Long-Term Interactive Memory},
  author={Wu, Di and He, Hongwei and Han, Wenhao and Pan, Weitao and Wang, William Yang},
  journal={arXiv preprint arXiv:2410.10813},
  year={2024}
}

@article{openai2023gpt4,
  title={{GPT}-4 Technical Report},
  author={OpenAI},
  journal={arXiv preprint arXiv:2303.08774},
  year={2023}
}

@article{openai2024gpt4o,
  title={{GPT}-4o System Card},
  author={Hurst, Aaron and others},
  journal={arXiv preprint arXiv:2410.21276},
  year={2024}
}

@article{anthropic2024claude,
  title={The {Claude} 3 Model Family: Opus, Sonnet, Haiku},
  author={Anthropic},
  journal={arXiv preprint arXiv:2407.01449},
  year={2024}
}

@article{dubey2024llama3,
  title={The {Llama} 3 Herd of Models},
  author={Dubey, Abhimanyu and Jauhri, Abhinav and Pandey, Abhinav and others},
  journal={arXiv preprint arXiv:2407.21783},
  year={2024}
}

@article{jiang2023mistral,
  title={Mistral {7B}},
  author={Jiang, Albert Q. and Sablayrolles, Alexandre and Mensch, Arthur and others},
  journal={arXiv preprint arXiv:2310.06825},
  year={2023}
}

@article{geminiteam2024gemini15,
  title={Gemini 1.5: Unlocking Multimodal Understanding Across Millions of Tokens of Context},
  author={{Gemini Team, Google}},
  journal={arXiv preprint arXiv:2403.05530},
  year={2024}
}

@article{sharma2023sycophancy,
  title={Towards Understanding Sycophancy in Language Models},
  author={Sharma, Mrinank and Tong, Meg and Korbak, Tomasz and others},
  journal={arXiv preprint arXiv:2310.13548},
  year={2023}
}

@article{fanous2025syceval,
  title={{SycEval}: Evaluating {LLM} Sycophancy},
  author={Fanous, Andrew and Goldberg, Jonathan and Agarwal, Aditya A. and others},
  journal={arXiv preprint arXiv:2502.08177},
  year={2025}
}

@article{ranaldi2023sycophantic,
  title={When Large Language Models Contradict Humans? Large Language Models' Sycophantic Behaviour},
  author={Ranaldi, Leonardo and Pucci, Giulia},
  journal={arXiv preprint arXiv:2311.09410},
  year={2023}
}

@article{hase2024belief,
  title={Fundamental Problems With Model Editing: How Should Rational Belief Revision Work in {LLMs}?},
  author={Hase, Peter and others},
  journal={arXiv preprint arXiv:2406.19354},
  year={2024}
}

@inproceedings{jang2024beliefr,
  title={Belief Revision: The Adaptability of Large Language Models Reasoning},
  author={Jang, Myeongjuk and others},
  booktitle={Proceedings of the 2024 Conference on Empirical Methods in Natural Language Processing},
  year={2024}
}

@inproceedings{lewis2020rag,
  title={Retrieval-Augmented Generation for Knowledge-Intensive {NLP} Tasks},
  author={Lewis, Patrick and Perez, Ethan and Piktus, Aleksandra and others},
  booktitle={Advances in Neural Information Processing Systems},
  volume={33},
  pages={9459--9474},
  year={2020}
}

@article{geng2025accumulating,
  title={Accumulating Context Changes the Beliefs of Language Models},
  author={Geng, Jiayi and Chen, Howard and Liu, Ryan and Horta Ribeiro, Manoel and Willer, Robb and Neubig, Graham and Griffiths, Thomas L.},
  journal={arXiv preprint arXiv:2511.01805},
  year={2025}
}

@article{dongre2025drift,
  title={Drift No More? {Context} Equilibria in Multi-Turn {LLM} Interactions},
  author={Dongre, Vardhan and Rossi, Ryan A. and Lai, Viet Dac and Yoon, David Seunghyun and Hakkani-Tur, Dilek and Bui, Trung},
  journal={arXiv preprint arXiv:2510.07777},
  year={2025}
}

@inproceedings{borah2025mind,
  title={Mind the ({Belief}) {Gap}: {Group} Identity in the World of {LLMs}},
  author={Borah, Angana and Houalla, Marwa and Mihalcea, Rada},
  booktitle={Findings of the Association for Computational Linguistics: ACL 2025},
  pages={18441--18463},
  address={Vienna, Austria},
  year={2025}
}

@article{cheng2025accommodation,
  title={Accommodation and Epistemic Vigilance: {A} Pragmatic Account of Why {LLMs} Fail to Challenge Harmful Beliefs},
  author={Cheng, Myra and others},
  journal={arXiv preprint arXiv:2601.04435},
  year={2025}
}

@article{methuku2025doppelgangers,
  title={Digital Doppelgangers: Ethical and Societal Implications of Pre-Mortem {AI} Clones},
  author={Methuku, Vijayalaxmi and Myakala, Praveen Kumar},
  journal={arXiv preprint arXiv:2502.21248},
  year={2025}
}

@article{myakala2025xai,
  title={The Human Factor in Explainable {AI} Frameworks for User Trust and Cognitive Alignment},
  author={Myakala, Praveen Kumar and Jonnalagadda, Anil Kumar and Bura, Chiranjeevi},
  journal={IARJSET},
  volume={12},
  number={1},
  year={2025},
  doi={10.17148/IARJSET.2025.12110}
}

@article{krippendorff2011,
  title={Computing Krippendorff's Alpha-Reliability},
  author={Krippendorff, Klaus},
  journal={Departmental Papers (ASC), University of Pennsylvania},
  year={2011}
}

@inproceedings{reimers2019sentence,
  title={Sentence-bert: Sentence embeddings using siamese bert-networks},
  author={Reimers, Nils and Gurevych, Iryna},
  booktitle={Proceedings of the 2019 conference on empirical methods in natural language processing and the 9th international joint conference on natural language processing (EMNLP-IJCNLP)},
  pages={3982--3992},
  year={2019}
}

@article{myakala2025fault,
  title={Fault-tolerant federated learning framework for edge devices in unstable networks},
  author={Myakala, Praveen Kumar and Agrawal, Manan},
  journal={Authorea Preprints},
  year={2025},
  publisher={Authorea}
}

@inproceedings{rudra2025composable,
  title={Composable AI Stack for Intelligent Agents: Modular Orchestration Using Context Routing, Memory, and Tools},
  author={Rudra, A and Agrawal, Manan},
  booktitle={2025 International Conference on Computer and Applications (ICCA)},
  pages={1--6},
  year={2025},
  organization={IEEE}
}

\end{document}